\documentclass[11pt]{article}
\usepackage{coling2020}
\usepackage{times}
\usepackage{url}
\usepackage{latexsym}

\usepackage{graphicx}
\usepackage{xcolor}
\usepackage{multirow}
\usepackage{bm}

\colingfinalcopy % Uncomment this line for the final submission

\title{Context in Informational Bias Detection}

\author{Esther van den Berg \\
  Leibniz ScienceCampus  \\
  Heidelberg/Mannheim, Germany \\
  {\tt berg@ids-mannheim.de} \\\And
  Katja Markert \\
  Institute of Computational Linguistics \\
  Heidelberg University, Germany \\
  {\tt markert@cl.uni-heidelberg.de} \\}

\date{}

\begin{document}
\maketitle
\begin{abstract}
   Informational bias is bias conveyed through sentences or clauses that provide tangential, speculative or background information that can sway readers' opinions towards entities. By nature, informational bias is context-dependent, but previous work on informational bias detection has not explored the role of context beyond the sentence. In this paper, we explore four kinds of context for informational bias in English news articles: neighboring sentences, the full article, articles on the same event from other news publishers, and articles from the same domain (but potentially different events). We find that integrating event context improves classification performance over a very strong baseline. In addition, we perform the first error analysis of models on this task. We find that the best-performing context-inclusive model outperforms the baseline on longer sentences, and sentences from politically centrist articles. 
\end{abstract}

%Furthermore, both models perform better on BASIL's Fox News instances than its New York Times or Huffington Post instances. Additionally, both models are more likely to predict that bias is present if the target sentences contains quotation marks. 

%
% The following footnote without marker is needed for the camera-ready
% version of the paper.
% Comment out the instructions (first text) and uncomment the 8 lines
% under "final paper" for your variant of English.
% 
\blfootnote{
    %
    % for review submission
    %
    %
    % % final paper: en-uk version 
    %
    % \hspace{-0.65cm}  % space normally used by the marker
    % This work is licensed under a Creative Commons 
    % Attribution 4.0 International Licence.
    % Licence details:
    % \url{http://creativecommons.org/licenses/by/4.0/}.
    % 
    % % final paper: en-us version 
    %
    \hspace{-0.65cm}  % space normally used by the marker
    This work is licensed under a Creative Commons 
    Attribution 4.0 International License.
    License details:
    \url{http://creativecommons.org/licenses/by/4.0/}.
}

\section{Introduction}
Informational bias is conveyed through sentences or clauses that provide tangential, speculative or background information that can sway readers' opinions towards entities \cite{basil2019}. 
A natural place to look for informational bias is in news texts, where journalists use background information to place newsworthy events in a broader context.
Examples of informational bias include quotations of opinions from third parties about the target entity, allusions to what may have motivated the target entity to act as they did, and mentions of previous statements and actions of the same entity.

What separates informational bias from other kinds of bias is that it can be expressed in a completely factual and neutral way. While some instances of bias are recognisable outside of their context  (e.g. quotations from third parties that contain opinions), others are mere statements of facts that do not raise suspicions of bias outside of their context. 
Consider example instance 1.3 in Table~\ref{tab:ex}. This sentence contains no subjective language. Seen on its own, it is simply stating a fact. However, human annotators judged that it is a case informational bias \cite{basil2019}. This is because this particular fact reflects positively on the target entity Mike Huckabee in the context of an announcement that he is running for president. Note that one can also imagine contexts where the implication is negative. An example would be a discussion of a disconnect between older Republican candidates and a new generation of more progressive voters. 
The fact that instances of informational bias can be superficially neutral and are  context-dependent makes informational bias detection an exceptionally challenging task, which furthermore has a short research history and few available relevant resources.

While previous work has performed informational bias detection on a token level and a sentence level, we are the first to  involve context beyond the sentence. 
We integrate four kinds of context: neighboring sentences (\textbf{direct textual context}), the full article (\textbf{article context}), articles on the same event (\textbf{event context}) and articles from the same domain (\textbf{domain context}).\footnote{Code and results available at: \url{github.com/vdenberg/context-in-informational-bias-detection}.} 
%Due to limited availability of data, {\color{red}{existing methods of integrating context}} struggle to improve performance over a strong sentence-level-only baseline. 
%KM: unclear what you mean here. What existing methods? Maybe be more clear here or just talk about the ones that work
Our model for leveraging event context significantly improves over the baseline.
In an error analysis, we find that the best-performing context-inclusive model outperforms the baseline on long sentences and sentences from politically centrist news articles.
%KM: should the subjective language part also go into the abstract?
\vspace{10pt}

Our contributions are:

\begin{enumerate}
    \item The first systematic study of the impact of including context in informational bias detection.
    \item The best-performing informational bias detection system for English thus far, tested on the BASIL corpus \cite{basil2019}.
    \item The first error analysis of informational bias detection systems, which identifies strengths and weaknesses of models with and without awareness of context.
\end{enumerate}

\begin{table}
\begin{center}
\begin{tabular}{|l|p{0.65\textwidth}|r|c|c|}
\hline \bf Idx & \bf Sentence & \bf Inf & \bf Src & \bf ID \\ \hline
\hline
%53hpo00 & 
%Mike Huckabee [\dots] announced Tuesday he'll run for the 2016 Republican presidential nomination. & HPO & 0 \\
1.1 & 
Former Arkansas Gov. Mike Huckabee announced Tuesday he is running for president [\dots]. &
0 & FOX & 53fox00  \\
\hline
1.2 & 
Mr. Huckabee opposes same-sex marriage, suggesting as recently as February that homosexuality is a lifestyle choice akin to drinking or swearing. &
1 & NYT & 53nyt10 \\
\hline
1.3 & 
He was the longest-serving Arkansas governor, from 1996 to 2007. &
1  & HPO  & 53hpo15 \\
\hline
%1.4 & 
%[Huckabee:] "A leader only starts a fight he's prepared to finish."  &
%1 & FOX & 53fox20 \\
%\hline

\hline
2.1  & 
“Trump says he wants to run the nation like he’s run his business,” Mr. Bloomberg said on Wednesday. & 
0 & NYT & 82nyt01 \\
\hline
2.2 & 
Bloomberg contrasted his business history with Trump's, saying "I’ve built a business, and I didn’t start it with a million-dollar check from my father."& 
1 & FOX & 82fox05 \\
\hline
2.3 & 
“But Trump’s business plan is a disaster in the making." & 
1 & HPO & 82hpo21 \\
\hline
%82nyt04 &They existed in different circles of New York’s ultrarich: Mr. Bloomberg is known as a generous philanthropist; Mr. Trump appears to have been flinty in his giving.& NYT & 1 \\ \hline

\hline
3.1 &
The states of Nebraska and Oklahoma filed a federal lawsuit in the U.S. Supreme Court Thursday [\dots]. &
0 & HPO & 21hpo00 \\
\hline
3.2 &
"While Colorado reaps millions from the sale of pot, Nebraska taxpayers have to bear the cost."   &
1 & FOX & 21fox09 \\
\hline
3.3 &
[Sheriff Adam Hayward of Deuel County, Neb.] has complained that marijuana arrests have strained his jail budget. &
1 & NYT & 21nyt09 \\
\hline

\end{tabular}
\end{center}
\caption{\label{tab:ex} \centering Example instances from three stories in the BASIL corpus \cite{basil2019} with their informational bias label (inf), news source (src) and BASIL ID. Note that Examples 2.3 and 3.2 
look superficially as if they contain explicit sentiment but are labeled as informational bias as they are only introduced as quotes into the news article.}
\end{table}

\section{Related work}
\textbf{Framing}. Informational bias can be considered a type of framing with a focus on entities. The framing of entities has been studied for the construction of the English BASIL corpus of lexical and informational bias \cite{basil2019}, which is the corpus our models are tested on. \newcite{basil2019} emphasize that, unlike more commonly studied kinds of bias, informational bias label assignments depend very heavily on context. The sentence-level and span-level BASIL annotations were thus provided by human annotators who saw sentences in their article context. However, the computational models in \newcite{basil2019}, which are based on BERT \cite{devlin2018bert}, only treat sentences in isolation.
Framing of entities is also studied by \newcite{card2016analyzing}, who examined how English-speaking news frames events through casts of characters, and \newcite{van2020doctor}, who studied the effect of naming and titling on the perception of entities in English and German.

Most framing research focuses not on entities but on the framing of topics and events. 
The study of topic framing in news has a long history in social science \cite{entman1993framing,berinsky2006making,baumgartner2008decline,gentzkow2010drives} and has begun to attract attention from the natural language processing community \cite{tsur2015frame,fulgoni2016empirical,field2018framing,baumer2015testing}.
For topic framing research there exists the Media Frames Corpus of news annotated for the framing of same-sex marriage, smoking, and immigration \cite{card2015media} and the Gun Violence Frame Corpus \cite{liu2019detecting} annotated for framing in news on gun violence.
Computational analysis and classification experiments have been done on framing in Russian news \cite{field2018framing}, on detecting frames in English headlines  \cite{liu2019detecting,chen2018learning}, and on detecting frames in a multi-label, multi-lingual setting \cite{akyurek2020multi}.

\textbf{Subjectivity.} Related work also includes work on implicit sentiment through syntactic structures \cite{greene2009more} and partisan phrases \cite{yano2010shedding}, work on explicit stance and subjective language \cite{recasens2013linguistic,pang2008opinion,wiebe2004learning,hube2019neural}, and work on the classification of documents or news outlets into leanings or ideologies \cite{iyyer2014political}. % (Entman, 2007; Gentzkow and Shapiro, 2006, 2010; Prat and Stro ̈mberg, 2013.
The difference between these various kinds of subjective language and informational bias lies in its exclusion of neutral and objective language. In framing research, any text that could lead an impartial third party to recognise a non-neutral viewpoint towards a topic or entity can be said to contain framing, even if it is objective and neutral in tone.

\textbf{Approaches.} Sentence-level classification tasks have seen great increases in performance through the use of pre-trained language models (PLMs) such as BERT  \cite{devlin2018bert} and RoBERTa \cite{liu2019roberta}.
By further pre-training RoBERTa on domain-specific and task-specific datasets, \newcite{gururangan2020don} made it possible to perform sentence-level classification using models that have been exposed to domain context.
In another line of work, several methods have been developed to allow PLM models to take larger sequences than sentences as their input \cite{pappagari2019hierarchical,adhikari2019docbert}. Of these, only one specifically performs sequential sentence classification (i.e. the task of providing labels for each of the sentences in the multi-sentence input) \cite{cohan2019pretrained}. There exist non-PLM approaches to sequential sentence classification as well. These consist of hierarchical sequence encoders with a final CRF layer \cite{dernoncourt2017pubmed,jin2018hierarchical} and a BiLSTM-based approach that contextualises Universal Sentence Encodings and also integrates information that is specific to the domain of movie plot synopses \cite{papalampidi2019movie}. 
None of these techniques have previously been applied to informational bias detection.

\section{Method}
We experiment with different kinds of context to assess which one or which ones are helpful for informational bias detection. We define four types of context: direct textual context, article context, event context and domain context.

%\vskip 0.5em

\textbf{Direct textual context.}
Direct textual context consists of the directly neighboring sentences around the target sentence. These may be helpful for disambiguating sentences with multiple possible interpretations, for noticing patterns in the type of content preceding and following instances of informational bias, and for noticing when a target sentence is part of a multi-sentence quote.

%\vskip 0.5em

\textbf{Article context.}
 Article context consists of the full news article that the target sentence appears in. The article may be helpful for e.g. establishing the topic of the target sentence, what type of article it is from, and whether the target sentence is an outlier compared to the rest of the article.

%\vskip 0.5em

\textbf{Event context}.
Event context consists of news articles that cover the same newsworthy event or topic as the article the target sentence appears in. They might appear in the same or different news outlets.
The possible benefit of access to Event context is that it is helpful for noticing when an article takes a unique stance on a topic or mentions information that is absent from other articles.

%\vskip 0.5em

\textbf{Domain context.} 
Domain context consists of the domain of news articles that the target sentence is a part of. Domain refers to a group of texts with shared lexical and structural properties, including register, topic and platform of publication. Domain context is therefore more general than event context. Following \newcite{gururangan2020don}, we consider the population of articles from which a corpus has been sampled a domain in its own right.
Possible benefits of domain awareness when detecting informational bias include the ability to notice typical journalistic strategies for framing entities without attracting accusations of bias, the ability to distinguish between news outlets differing styles and ideologies, and increased awareness of domain-specific connotations of words and phrases (e.g. ``leading in the polls'' or ``declined to comment'').

\begin{figure*}[h!]
  \centering
  \includegraphics[width=.98\textwidth]{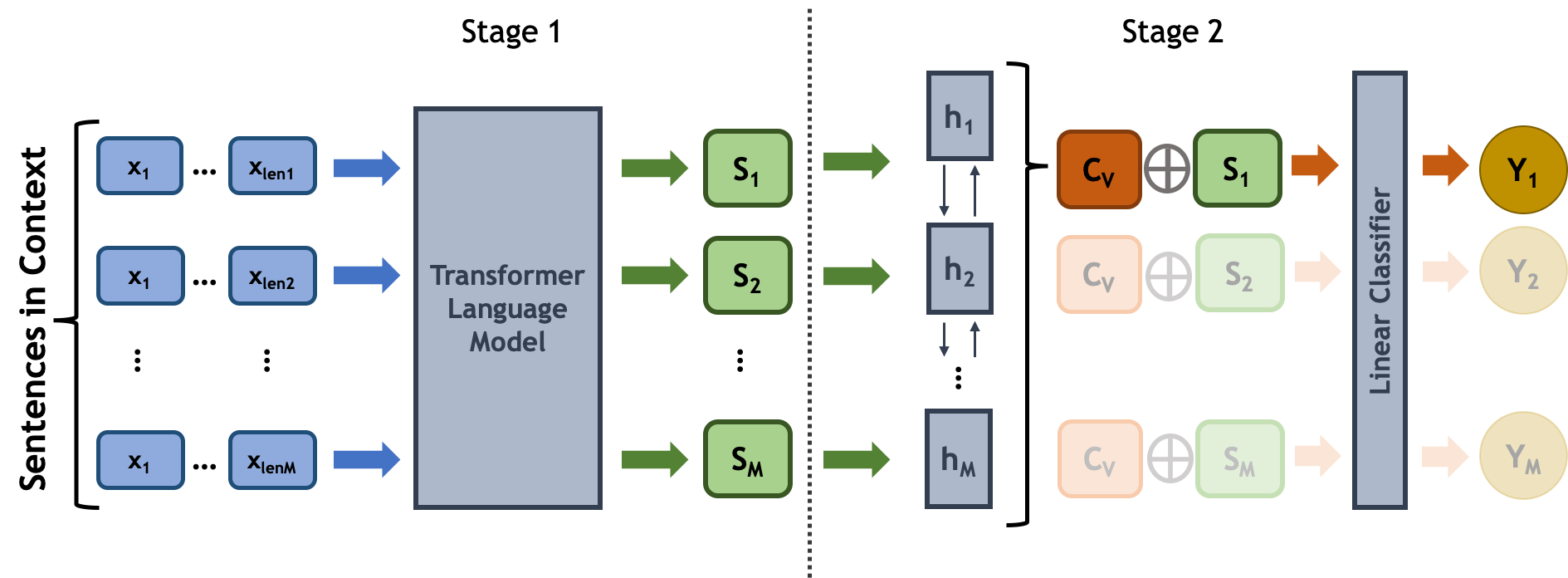}
  \caption{\label{fig:cim} Diagram of the Context-Inclusive Model. In Stage 1, Sentence embeddings are obtained by encoding word sequences using a pre-trained language model. In Stage 2,  Sequences of sentences from a pre-determined context (article or event) are encoded by as many BiLSTMs as there are documents (only one shown in diagram). The resulting context representation is concatenated and classified together with the target sentence representation to obtain a sentence-level prediction.}
\end{figure*}
\subsection{Data \& Task}

We use the currently only existing dataset with annotations of informational bias: the aforementioned BASIL dataset. 
This corpus contains 100 triples of news articles by Fox News (FOX), the New York Times (NYT) and  the Huffington Post (HPO), each covering the same event. 
The dataset consists of 7,984 sentences. During pre-processing, we remove 7 empty instances from the corpus with a sentence length of zero, leaving 7,977 instances. The corpus provides span-level annotations which can be used for token classification, or for binary sentence classification by converting them into sentence labels. A sentence is labeled as biased if it contains at least one span with bias. The total number of sentences with informational bias amounts to 1,221.
Some example sentences from  this corpus are given in Table~\ref{tab:ex}. 

The context-inclusion experiments in this work are only performed on sentence classification, as several of the proposed models are more suited for this task.
For completeness, we also provide baseline results for token classification.

\subsection{Approaches}\label{approaches}

\textbf{Baseline.} 
We compare our context-inclusive models to \textbf{BERT} \cite{devlin2018bert} and \textbf{RoBERTa}  \cite{liu2019roberta} for binary sentence classification. These are both powerful transformer language models pre-trained on large amounts of data and proven to be effective for low-resource tasks.
They take as input single sentences and output labels. They thus do not consider the context  which sentences appear in, but are optimised to make excellent use of any cues contained within the sentence.

\textbf{Direct textual context.} To involve direct textual context, we use a Windowed Sequential Sentence Classification method (\textbf{WinSSC}).
Like \newcite{cohan2019pretrained}'s method of using pre-trained language models for sequential sentence classification, WinSSC takes multiple sentences as its input sequence, generates embeddings for the separator tokens in the sequence, and classifies these embeddings with a linear layer that outputs as many labels as there are sentences in the input sequence.
Prior to embedding, sequences are book-ended with the last sentence from the previous sequence, and the first sentence of the next sequence.
These book-ends, which are ignored during evaluation, ensure that each sentence in the sequence has context at both ends, thus mitigating loss of information along the edges of sequences.
This is important when segmenting news articles as they tend to be long enough to require segmentation into several sections to avoid memory problems.
We experiment with sections of 5 and 10 sentences to assess the effect of changing section sizes, and we compare our WinSSC method to the non-windowed SSC method from \newcite{cohan2019pretrained}. 
\vspace{5pt}

\textbf{Article context and event context.} We integrate article context and event context by means of an Article Context-Inclusive Model (\textbf{ArtCIM}) and Event Context-Inclusive Model (\textbf{EvCIM}). 
Inspired by the Context Aware Model in \newcite{papalampidi2019movie}, the Context-Inclusive Model uses a Bidirectional Long Short-Term Memory (BiLSTM; Hochreiter and Schmidhuber 1997) to encode news documents.
In the case of ArtCIM, a single BiLSTM encodes the article. In the case of EvCIM, three BiLSTMs encode each document in the triple of Fox News, New York Times and Huffington Post articles on the same event. 
Sentence representations for the target sentence as well as for the input to the BiLSTMs are obtained by taking the average of the last four layers of fine-tuned base RoBERTa (Figure~\ref{fig:cim}, Stage 1). 
We found this to be more effective than other kinds of pooling, and also more effective than sentence USE embeddings \cite{cer2018universal} or Sentence-Bert embeddings \cite{reimers2019sentence}.
BiLSTMs then encode a context representation of the article the target sentence appears in (ArtCIM), or of each of the three articles on the same event (EvCIM).
At the final stage, the encodings of the target sentence and the context documents are concatenated and passed to a linear classifier (Figure~\ref{fig:cim}, Stage 2).
Classification is thus based both on the content of the target sentence, which the baseline captures very well, and the article or event context, which the baseline has no access to.

%\item[Averaged Input] The BERT language model \cite{devlin2018bert} uses a transformer to contextualise tokens. The average final hidden state of tokens in a sequence is a recommended representation of that sequence according to BERT documentation \cite{reimers2019sentence}. 
%\item[Crosslayer Input] In this setting, we take the average across the last four layers of a language model and use the result as a representation of the sentence.

\textbf{Domain context.} To integrate domain context, we apply domain-adapted RoBERTa from \newcite{gururangan2020don} that has been trained on news data (\textbf{DAPT}), a task-adapted version of RoBERTa that has been trained on the BASIL data (\textbf{TAPT}), or both (\textbf{DAPT-TAPT}).
Additionally, we experiment with including domain context by concatenating an embedding representing the source of an article (FOX, NYT or HPO) as a feature in the CIM setting (at Stage 2 in Figure~\ref{fig:cim}) (\textbf{ArtCIM*} and \textbf{EvCIM*})\footnote{We experimented with combining ArtCIM and EvCIM with domain-adapted and task-adapted RoBERTa, but combining the two lowered performance.}.

\section{Results}
\begin{table*}
\begin{center}
\begin{tabular}{|l|l|l|ccc|}
\hline \bf Task & \bf Set-Up & \bf Model & \bf Precision & \bf Recall & \bf F1-score \\ 
\hline
\hline

\multirow{4}{*}{Token}  & \multirow{2}{*}{Sentence Split} 
& \newcite{basil2019}
& $25.56$      & $14.78$      & $18.71$  \\
&& BERT
& $12.42 \pm 1.31$      & $28.31 \pm 3.18$      & $17.23 \pm 1.61$  \\ %old_eval
&& RoBERTa
& $36.10 \pm 4.51$      & $32.41 \pm 2.92$    & $34.03 \pm 2.81$   \\ %new_eval

\cline{2-6}

                        & \multirow{2}{*}{Story Split} 
& BERT
& $12.85 \pm 1.06$      & $22.12 \pm 1.60$    & $14.60 \pm 0.91$  \\ %old_eval
&& RoBERTa
& $32.44 \pm 2.04$      & $27.73 \pm 1.54$     & $29.86 \pm 1.25$  \\  %new_eval
\hline

\hline

\multirow{4}{*}{Sentence} & \multirow{2}{*}{Sentence Split}    
& \newcite{basil2019}
&  $43.87$         & $42.91$    & $43.27$  \\
&& BERT  
&  $46.44 \pm 2.51$         & $33.0 \pm 7.21$    & $ 38.26 \pm 5.29$  \\ %new_eval
&& RoBERTa   
& $47.55 \pm 2.92$  & $52.67 \pm 6.41$   & $49.89 \pm 4.06$ \\ %new_eval
\cline{2-6}
  
& \multirow{2}{*}{Story Split}    
& BERT  
&  $38.96 \pm 5.55$         & $35.79 \pm 2.15$    & $37.00 \pm 2.11$  \\ %new_eval
&& RoBERTa  
& $43.12 \pm 1.03$  & $41.29 \pm 1.37$ & $42.16 \pm 0.30$  \\ %new_eval
\hline
\end{tabular}
\end{center}
\caption{\label{baselines} \centering  Baseline performance  without context
for token and sentence classification. Sentences are either   divided across training and non-training sets by sentence or by story. We report standard deviations across 5 seeds for our models. \newcite{basil2019} report a minimum standard deviation of 3.36 and maximum of 12.44 for theirs.}
\end{table*}
\begin{table}
\begin{center}
\begin{tabular}{|l|r|r|r|}
\hline \bf Model & \bf Precision & \bf Recall & \bf F1-score \\ \hline
RoBERTa 
& $43.12 \pm 1.03$  & $41.29 \pm 1.37$ & $42.16 \pm 0.30$  \\ %new_eval
\hline
SSC-5 
& $41.90 \pm 1.00$  & $36.16 \pm 1.13$ & $38.19 \pm 0.98$   \\
SSC-10 
& $43.84 \pm 1.64$  & $34.88 \pm 0.71$ & $38.22 \pm 1.11$  \\

WinSSC-5 
& $42.28 \pm 0.99$  & $36.94 \pm 0.88$ & $38.67 \pm 0.82$ \\
WinSSC-10
& $43.20 \pm 1.37$  & $35.12 \pm 2.41$ & $37.44 \pm 0.79$ \\
\hline
\end{tabular}
\end{center}
\caption{\label{SSC} \centering Results of integrating direct textual context with a Sequential Sentence Classifier without (SSC \cite{cohan2019pretrained}) or with a window (WinSSC) and a maximum sequence length of 5 or 10.}
\end{table}
%KM: precision/recall here are the only ones without stddeviation?

\subsection{Set-Up}

Previous work on the BASIL corpus has split data by dividing sentences across a training, development and test set \cite{basil2019}. 
This type of split isolates target sentences from other sentences in the same article, and from other articles covering the same event. 
Distributing sentences across set types in this way is contrary to the goal of this work, which is to consider sentences within their context. 
In addition, distributing sentences from the same article across training and test data can be considered a type of leakage, as knowing of some sentences in an article that they are biased might help recognise similar sentences from the same article or another article on the same topic.
Our setting, in which triples of articles appear either during training or during testing but not both, resembles a more realistic setting, where the hypothetical user of an information bias annotation system wants to identify bias in new articles on new events.

We use the split with sentences distributed across train and non-train sections - the \textbf{Sentence split} - only to report baseline results for the purpose of consistency and comparability to \newcite{basil2019}. 
This split consists of 7,123 training instances, 408 development instances and 404 test instances.
To test context-inclusive methods, we use a 10-fold cross-validation setting where \textit{stories} (triples of articles) never appear in both a train and non-train section. 
%{\color{red} Mention ensemble methods here? Which citations?}
Sizes of folds in this \textbf{Story split} vary slightly because of variation in the length of articles.
Each consists of around 6,400 sentences designated for training, 780 for development and 790 for testing.
All methods are tested 5 times with a different random seed.  We report precision, recall and balanced F-measure (with standard deviation across seeds) for the positive (biased) class and test significance of differences in performance with an independent t-test.
Further training details are provided in Appendix A.

%\footnote{\url{https:dummy_url}}. %~\ref{sec:appendixA}

\subsection{Baseline}
To establish a baseline that classifies sentences in isolation we fine-tune the language models BERT \cite{devlin2018bert} and RoBERTa \cite{liu2019roberta} on the BASIL corpus.
For comparison with \newcite{basil2019}, we fine-tune both to perform token classification and sentence classification and on both the Sentence split and Story split. 

In line with the prediction that the Sentence split introduces leakage from test into training data, performance is several F1-score points higher on the Sentence split than on the Story split in all settings (Table~\ref{baselines}). 
The difference is largest for sentence classification with RoBERTa (F1=$49.89$ on the Sentence split and F1=$42.16$ on the Story split).

In line with observations in \newcite{basil2019}, we also observe that performance is lower for token classification than for sentence classification (F1=$29.86$ versus F1=$42.16$ (RoBERTa on the Story split)).

We observe large improvements in performance of RoBERTa over BERT in all settings.
Best performance on sentence classification was reported to be F1=$43.27$ on a Sentence split in \newcite{basil2019} by BERT.
In our set-up using our seeds, BERT performance stands at F1=$38.26$ on the Sentence split, whereas RoBERTa's sentence classification performance on the Sentence split is $49.89$.
On the Story split the difference is also large: from F1=$37$ by BERT to F1=$42.16$ by RoBERTa.

\subsection{Direct Textual Context}
We experiment with integrating direct textual context by comparing two methods of sequential sentence classification (SSC and the novel WinSSC) to the best performing baseline sentence classifier. 
We find performance decreases when direct context is introduced in this manner (Table~\ref{SSC}).
Increasing the length of the sequence from 5 to 10 does not aid performance of either the non-windowed SSC or the WinSSC model (F1=$38.19$ to F1=$38.22$ and F1=$38.67$ to F1=$37.44$). 
It is likely that data sparsity is at fault here.
When performing 10-fold cross-validation with the maximum sequence length set to 5, the number of sequences for training averages around 1654 per iteration. With the maximum sequence length set to 10, this drops further to 856 sequences. 
This is likely too small a number of sequences for the models to generalize.

\subsection{Article Context \& Event Context}
We use the Context-Inclusive Model to perform classification based on encodings of the target sentence and either only the news article the target sentence appears in (ArtCIM) or each member of the triple of coverage on the same event (EvCIM). As an additional event context experiment, we provide ArtCIM and EvCIM with a representation of the news source (ArtCIM* and EvCIM*). 
Each of these models performs at least as well as the baseline. 
While RoBERTa achieves best precision, all CIM models achieve much higher recall.
In terms of F1-score, EvCIM performs significantly better than the baseline ($p<.001$), as does EvCIM* ($p=.004$).
%EM: test mentioned ($t(4)=12.8, p<.001$)($t(4)=-3.938, p=.004$)

\subsection{Domain Context}
We test the performance of three adaptations of RoBERTa: domain-adapted to news (DAPT), task-adapted for informational bias detection on the BASIL corpus (TAPT) and domain-and-task-adapted (DAPT-TAPT).
The DAPT and DAPT-TAPT model do not outperform the baseline, but the task-adapted model does (Table~\ref{dapt}), although not significantly.
These results echo the finding in \newcite{gururangan2020don} that domain-adaptation and domain-and-task-adaptation are not as helpful in the news domain as in other domains.
RoBERTa has likely seen a sufficient amount of news domain training data already \cite{liu2019roberta}, making any further domain-specific pre-training marginally helpful as long as the domain is kept as general as news.

\begin{table}
\begin{center}
\begin{tabular}{|l|r|r|r|}
\hline \bf Model & \bf Precision & \bf Recall & \bf F1-score \\ 
\hline
RoBERTa
& $43.12 \pm 1.03$  & $41.29 \pm 1.37$ & $42.16 \pm 0.30$  \\ %new_eval
\hline
 ArtCIM 
& $38.81 \pm 0.93$   & $47.78 \pm 1.82$    & $42.80 \pm 0.55$   \\%old_eval 
ArtCIM*
& $39.08 \pm 0.53$   & $45.18 \pm 1.25$   & $42.31 \pm 0.63$ \\  %new_eval
EvCIM 
& $39.72 \pm 0.59$   & $49.60 \pm 1.20$   & $44.10 \pm 0.15$ $^{\dagger}$  \\  %new_eval
EvCIM* 
& $39.76 \pm 1.50$   & $46.88 \pm 2.42$   & $42.96 \pm 0.34$ $^{\dagger}$  \\ %new_eval
\hline
\end{tabular}
\end{center}
\caption{\label{CIM} \centering Results of article and event context with a Context-Inclusive Model with (*) or without news source as an added feature. A dagger indicates a significant improvement over the baseline.}
\end{table}

%TODO DAGGER IN SUPERSCRIPT AND MENTION THAT IT IS SIGNIFICANCE 
\begin{table}
\begin{center}
\begin{tabular}{|l|r|r|r|}
\hline \bf Model & \bf Precision & \bf Recall & \bf F1-score \\ 
\hline
RoBERTa 
& $43.12 \pm 1.03$  & $41.29 \pm 1.37$ & $42.16 \pm 0.30$  \\ %new_eval
\hline
DAPT 
& $46.87 \pm 1.32$   & $36.45 \pm 1.27$  &  $40.97 \pm 0.32$  \\ %new_eval
%HYPER-TAPT
%& $45.74 \pm 1.4$   & $40.38 \pm 0.94$  &  $42.13 \pm 1.16$   & $4/5$ \\
%HYPER-DAPTTAPT
%& $45.88 \pm 1.5$   & $37.32 \pm 1.29$  &  $40.25 \pm 0.45$  & $4/5$  \\
TAPT
& $46.49 \pm 1.74$  & $40.28 \pm 1.91$ & $43.12 \pm 1.07$ \\ %new_eval
DAPT-TAPT
& $46.69 \pm 1.50$  & $37.41 \pm 2.37$ & $41.47 \pm 1.00$  \\ %new_eval
\hline
\end{tabular}
\end{center}
\caption{\label{dapt} \centering  Results of domain-adapted, task-adapted and domain-and-task-adapted RoBERTa.}
\end{table}

\section{Error Analysis and Discussion}
To investigate whether the best-performing context-inclusive model (EvCIM) improves over the baseline by, in fact, leveraging context, we analyze dependence of performance on factors that we suspect influence the need for context when detecting informational bias. The factors that we consider are sentence length, the presence of quotes, the political leaning of the source article and the presence of subjective language. 

%We suspect that sentence length influences the complexity of the target instance, and that less context is needed to classify short, simple sentences. 
%We consider quotes because it has been shown that quoting patterns introduce bias in news \cite{niculae2015quotus} and informational bias in particular \cite{basil2019}, as they maintain an air of neutrality on the part of the journalist. 
%We suspect informational bias is less discreet and easier to classify when they are from sources with an identifiable political leaning.
%We measure two aspects of neutrality. The first is the political leaning of the news source that published the article. The second is the political leaning of the article the instance is from. %The third is the degree of lexical bias (i.e. sentiment-bearing words) in the article.

%Concretely, we expect that less context is needed and therefore \textit{fewer} gains of EvCIM over the baseline can be expected for sentences with the following characteristics:
Concretely, we expect that more context is needed and therefore \textit{higher} gains of EvCIM over the baseline can be expected for sentences with the following characteristics:

\begin{enumerate}
%\item They are {\color{red}{short}}, and more likely to contain a simple message that is easier to classify (e.g. 2.3 in Table~\ref{tab:ex}).
%KM: we talked about this before and it still reads counter-intuitive...
\item They are long, and consequently contain a more complex message that requires more knowledge to interpret (e.g. 2.2 in Table~\ref{tab:ex}, in contrast to 2.3).
%\item They are a quote or contain quotes (e.g. 2.1 and 2.3 in Table~\ref{tab:ex}). Quotes have been shown to introduce bias in news \cite{niculae2015quotus} and informational bias in particular \cite{basil2019}, as they maintain an air of neutrality on the part of the journalist. A simple model without context might just pick up on quotation marks as a clue for informational bias.
\item They are not a quote (e.g. 3.3 in Table~\ref{tab:ex}, in contrast to 3.2). Quotes have been shown to introduce bias in news \cite{niculae2015quotus} and informational bias in particular \cite{basil2019}, as they maintain an air of neutrality on the part of the journalist. A simple model without context
might just pick up on quotation marks as a clue for informational bias.
%\item They are from an article with a non-centrist political leaning (e.g. 2.3 and 3.2 in Table~\ref{tab:ex}).
\item They are from an article with a centrist political leaning (e.g. 1.2 and 3.3 in Table~\ref{tab:ex}).
%\item They do not contain directly subjective language (e.g. 3.3 in Table~\ref{tab:ex}).
\item They do not contain any subjective language (e.g. 1.3 in Table~\ref{tab:ex}, in contrast to 2.3).
\end{enumerate}

%By contrast, we predict \textit{fewer} gains in performance by EvCIM on short sentences, sentences that are quotes, sentences from non-centrist articles and sentences with subjective language. 

\subsection{Sentence Length}
\begin{table}
\begin{center}
\begin{tabular}{|r|r|r|r|r|}
\hline  
\bf Q & \bf N &\bf  Bias &\bf RoBERTa &\bf EvCIM \\ 
\hline
0-18     &  $2020$ &  $13.07\%$ &  $44.50$ &  $44.67$   \\
19-27  &  $2184$ &  $13.69\%$ &  $40.00$ &  $40.78$  \\
28-36  &  $1821$ &  $15.82\%$ &  $39.61$ &  $42.14\dagger$  \\
37-109 &  $1952$ &  $18.95\%$ &  $44.10$ &  $47.79\dagger$ \\
\hline
All        &  $7977$ &  $15.31\%$ &  $42.16$ &  $44.10\dagger$ \\
\hline
\end{tabular}
\end{center}
\caption{\label{sent_length} \centering  Performance (F1) by sentence length (quartiles) of RoBERTa and EvCIM. The first column shows the number of tokens per sentence in each quartile, the second column the number of sentences in each quartile and the third one the percentage of biased sentences per quartile.  A dagger indicates a significant improvement over the baseline.}
\end{table}

We examine whether EvCIM outperforms the baseline on longer sentences by partitioning data into bins corresponding to quartiles of sentence length (number of tokens). 
We then compare the F1-score computed on predictions by RoBERTa and EvCIM.
We observe that, as suspected, EvCIM does not outperform the baseline significantly on the shorter sentences, i.e. on the first and second quartile (Table~\ref{sent_length}).
On the longer sentences in the third and fourth quartile, however, EvCIM outperforms the baseline significantly (from F1=$39.61$ to F1=$42.14$ and from F1=$44.10$ to F1=$47.79$).

%We also examine whether EvCIM outperforms the baseline on longer spans of informational bias by partitioning data into bins corresponding to different proportions of biased tokens to unbiased tokens. 
%We then compare the f1-score computed on predictions by RoBERTa and EvCIM on the entire dataset (Table~\ref{span_length}). 

\subsection{Quotes}
\begin{table}
\centering
\begin{tabular}{|l|l|l|r|r|}
\hline  
\bf  In Quote  & \bf N & \bf RoBERTa  & \bf EvCIM \\ 
\hline
No        &   $634$ &   $28.33$ &   $39.05\dagger$ \\
Yes       &   $587$ &  $55.30$ &   $60.99\dagger$ \\
\hline
All Biased    &  $1221$ &  $41.29$ &   $49.60\dagger$ \\
\hline
\end{tabular}
\caption{\label{inf_quote} \centering  Recall of bias inside or outside of a quote of RoBERTa and EvCIM (biased instances only).}
\end{table}

Quoting patterns have been shown to introduce bias in news \cite{niculae2015quotus} and informational bias in particular \cite{basil2019}, by introducing opinions through a third party proxy.
We predict that neural approaches notice this relationship and rely on it to some degree to make their predictions. %, possibly at the expense of precision.

The BASIL corpus contains annotations that specify for each instance of informational bias whether it is part of a quote or not.
We can therefore analyze differences in recall of informational bias inside and outside of quotes (Table~\ref{inf_quote}).
Table~\ref{inf_quote} shows that both models have considerably better recall of informational bias in quotes.
We predicted that the baseline would have an easier time with quotes, and that the gains of EvCIM with respect to the baseline would be higher on non-quotes.
EvCIM outperforms the baseline in terms of recall by a higher margin on non-quotes than on quotes, but it outperforms it significantly on both.

%Because quotes are denoted orthographically with double quotation marks, we ask whether models rely on the presence of these marks when predicting bias.
%To test this we automatically annotate the data for the presence of quotation marks and look for over-representation of sentences with quotation marks among RoBERTa's and CIM's false positives Table~\ref{quote}.
%Note that quotes can be several sentences long, in which case only the first and last sentence contain quotation marks. 
%The true number of sentences in quotes is thus larger than what our automated annotation returns.

%Table~\ref{quote} shows that both models' false positives contain a high proportion of quotation marks, but that the proportion in the false positives of CIM with coverage context is lower.
%This suggests coverage context may have helped CIM to not rely as heavily on the orthographic cue that quotation marks provide.

\subsection{Political Leaning of the Source}
\begin{table}
\parbox{.48\linewidth}{
\begin{center}
\begin{tabular}{|l|r|r|r|r|}
\hline  
\bf Pub & \bf N & \bf \%Bias & \bf RoB & \bf EvCIM \\ 
\hline
FOX    &  $2633$ &  $15.65\%$ &  $46.12$ &  $47.31\dagger$ \\
NYT    &  $3048$ &  $14.93\%$ &  $39.78$ &  $43.01\dagger$ \\
HPO    &  $2296$ &  $15.42\%$ &  $40.52$ &  $41.82\dagger$ \\
\hline
All    &  $7977$ &  $15.31\%$ &     $42.16$ &      $44.10\dagger$ \\
\hline
\end{tabular}
\end{center}
\caption{\label{source} \centering  Performance (F1) by publisher of RoBERTa and EvCIM.}
}
\hfill
\parbox{.48\linewidth}{
\begin{center}
\begin{tabular}{|l|r|r|r|r|}
\hline  
\bf Lean & \bf N &  \bf \%Bias & \bf RoB & \bf EvCIM \\ 
\hline
Right  &  $2010$ &  $15.82\%$ &  $43.22$ &  $43.92$ \\
Center &  $3660$ &  $14.07\%$ &  $42.66$ &  $45.40\dagger$ \\
Left   &  $2307$ &  $16.82\%$ &  $40.68$ &  $42.58\dagger$ \\
\hline
All    &  $7977$ &  $15.31\%$ &  $42.16$ &  $44.10\dagger$ \\
\hline
\end{tabular}
\end{center}
\caption{\label{lean} \centering  Performance (F1) by article leaning (left, center, right) of RoBERTa and EvCIM.}
}
\end{table}

To examine whether bias is easier to detect in newspapers and articles with a more pronounced political leaning, we first compare performance on the different news publishers represented in the BASIL corpus. According to \newcite{budak2016fair}, FOX is strongly right-leaning, NYT slightly left-leaning, and HPO strongly left-leaning, meaning NYT is the most centrist of the three.
We observe that EvCIM significantly outperforms the baseline for all publishers, and that both models perform much better on Fox News articles compared to the other publishers (Table~\ref{source}).
\newcite{basil2019} have stated that there are differences in the polarity and target of biased sentences in the three news sources included in the BASIL corpus.
The RoBERTa and EvCIM systems may be capitalizing on these and other differences to make better predictions for Fox News articles.

In addition to providing the news source for each article, the BASIL corpus also provides article-level annotations of the political leaning of the article as determined by a human annotator. 
These annotations show that although publishers lean towards a certain side of the political spectrum, they also each publish a large number of centrist articles (Table~\ref{publisherleaning}).
When using these human annotations of leaning to compare performance on centrist and non-centrist articles, we find that EvCIM does not outperform the baseline significantly on right-leaning articles, but does outperform it significantly on left-leaning articles (from $F1=40.68$ to $F1=42.58$), and especially on centrist articles (from $F1=42.66$ to $F1=45.50$), supporting the notion that classification of these articles benefits more from access to event context.
%If we treat publisher and leaning as ordinal variables with the levels FOX ($-1$), NYT ($0$) and HPO ($1$) and Right ($-1$), Center ($0$) and Left ($1$) respectively, we can conduct a spearman's test of correlation which shows only a moderate correlation between the two variables ($\rho=.37$).

\begin{table}
\centering
\begin{tabular}{|l|r|r|r|r|}
\hline  
\bf Pub   & \bf Right & \bf Center  & \bf Left & \bf All  \\ 
\hline
FOX &   $50$ &   $38$ &   $12$ & $100$ \\
NYT &   $15$ &  $54$ &   $31$ & $100$ \\
HPO &   $10$ &  $52$ &   $38$ & $100$ \\
\hline
All &   $75$ &  $144$ &   $81$ & $300$ \\
\hline
\end{tabular}
\caption{\label{publisherleaning} \centering  Number of right-leaning, centrist and left-leaning articles from each news publisher.}
\end{table}

%EvCIM performs best on centrist articles, and its performance exceeds baseline performance for centrist articles (from F1=$42.66$ to ) for right-leaning (from  to F1=$43.92$) and left-leaning ones (from F1=$40.68$ to F1=$42.58$), confirming that EvCIM's gains over the baseline are higher on sentences from more neutral articles. 

\subsection{Subjective language}
\begin{table}
\parbox{.49\linewidth}{
\begin{center}
\begin{tabular}{|r|r|r|r|r|}
\hline  
\bf Lex & \bf N &\bf  Bias &\bf RoBERTa &\bf EvCIM \\ 
\hline
Yes       &   $448$ &   $9.82\%$ &  $27.39$ &  $25.30\ddagger$ \\
No        &  $7529$ &  $15.63\%$ &  $43.42$ &  $45.79\dagger$ \\
\hline
All       &  $7977$ &  $15.31\%$ & $42.16$ &  $44.10\dagger$ \\
\hline
\end{tabular}
\end{center}
\caption{\label{lex} \centering  Performance (F1) on items with and without lexical bias of RoBERTa and EvCIM. A double dagger indicates significantly worse performance than the baseline.}
}
\hfill
\parbox{.50\linewidth}{
\begin{center}
\begin{tabular}{|r|r|r|r|r|}
\hline  
\bf Subj & \bf N &\bf  Bias &\bf RoBERTa &\bf EvCIM \\ 
\hline
Yes  &  $2415$ &  $18.92\%$ &  $43.01$ &  $45.27\dagger$ \\
No   &  $5562$ &  $13.74\%$ &  $40.62$ &  $41.97\dagger$ \\
\hline
All  &  $7977$ &  $15.31\%$ &  $42.16$ &  $44.10\dagger$ \\
\hline
\end{tabular}
\end{center}
\caption{\label{subj} \centering  Performance (F1) on items with and without subjectivity clues of RoBERTa and EvCIM.}
}
\end{table}

We suspect that EvCIM will make larger gains on sentences without subjective language. 
The BASIL corpus contains annotations of lexical bias, i.e. bias through word choice, that can be used to investigate whether informational bias detection is helped by the presence of lexical bias. 
According to the BASIL annotation protocol, sentences contain lexical bias if the annotator found their opinion to be swayed by the choice of words. 
According to the annotators, this was the case for only 448 sentences, and informational bias was less likely to occur in sentences with lexical bias ($9.82\%$) than sentences without lexical bias ($15.63\%$) (Table~\ref{lex}). 
For comparison, we also compute the number of sentences that contain at least one strongly subjective  clue from the MPQA Subjectivity Lexicon \cite{wilson2005recognizing}.
We find that this number is higher: 2415 instances, and informational bias was \textit{more} likely to occur in sentences with subjectivity ($18.92\%$) than sentences without subjectivity ($13.74\%$) (Table~\ref{subj}). 
We suspect that the latter numbers are a more realistic assessment of the amount of subjective language in the BASIL corpus.

When comparing performance on instances with and without subjective language using the subjectivity lexicon, we observe only a small difference in improvement, with EvCIM outperforming the baseline more on items without subjective language (from F1=$40.62$ to F1=$41.97$) than with subjective language (from F1=$43.01$ to F1=$45.27$).
%KM: we need to talk about what higher means. Normally this should be reducton in error rate or improvement
%considered in comparison to original and then these are basically the same. We should discuss this friday

%\section{Discussion}
%\input{sections/discussion.tex}

\section{Conclusion}
We explore the impact of including four kinds of context in informational bias detection.
We integrate direct textual context, article context, context from other articles on the same event (event context), and domain context into sentence classification methods and test performance on the BASIL corpus of informational bias. 
We find that direct textual context and domain context are difficult to integrate in a way that boosts performance beyond the strong RoBERTa baseline. 
Our proposed Context-Inclusive Model, however, outperforms RoBERTa signficantly when using event context (EvCIM). 
Error analysis shows that EvCIM performs better than the baseline on longer sentences, and sentences from politically centrist articles.
Furthermore, both models perform better on BASIL's Fox News instances than its New York Times or Huffington Post instances, and both models are better at recognising bias in quotes. 

% both struggle with detecting positive informational bias.
% RoBERTa also struggles with articles that do not prominently feature well-known entities, a weakness which CIM does not appear to suffer from.

%The influence of news source origin on model performance suggests categorical differences in the way news sources use background information to influence opinion.
Future work could explore domain-adaptation to unlabeled data from the same population of articles that the BASIL corpus was drawn from. Given the differences in performance on Fox News Articles compared to other sources, domain-adaptation to specific sources is also a promising avenue. In addition, future datasets may need to ensure a balance of sources that represent different layers and sections of society.
Future work could also extend context-inclusion experiments to token classification.

\section*{Acknowledgements}
This research is funded by the Leibniz Science-Campus \textit{Empirical Linguistics \& Computational Language Modeling}, supported by Leibniz Association grant no. SAS2015-IDS-LWC and by the Ministry of Science, Research, and Art of Baden-W\"urttemberg.

\bibliographystyle{coling}
\bibliography{coling2020}

\begin{thebibliography}{}

\bibitem[\protect\citename{Adhikari \bgroup et al.\egroup
  }2019]{adhikari2019docbert}
Ashutosh Adhikari, Achyudh Ram, Raphael Tang, and Jimmy Lin.
\newblock 2019.
\newblock Docbert: Bert for document classification.
\newblock {\em arXiv preprint arXiv:1904.08398}.

\bibitem[\protect\citename{Aky{\"u}rek \bgroup et al.\egroup
  }2020]{akyurek2020multi}
Afra~Feyza Aky{\"u}rek, Lei Guo, Randa Elanwar, Prakash Ishwar, Margrit Betke,
  and Derry~Tanti Wijaya.
\newblock 2020.
\newblock Multi-label and multilingual news framing analysis.
\newblock In {\em Proceedings of the 58th Annual Meeting of the Association for
  Computational Linguistics}, pages 8614--8624.

\bibitem[\protect\citename{Baumer \bgroup et al.\egroup
  }2015]{baumer2015testing}
Eric Baumer, Elisha Elovic, Ying Qin, Francesca Polletta, and Geri Gay.
\newblock 2015.
\newblock Testing and comparing computational approaches for identifying the
  language of framing in political news.
\newblock In {\em Proceedings of the 2015 Conference of the North American
  Chapter of the Association for Computational Linguistics: Human Language
  Technologies}, pages 1472--1482.

\bibitem[\protect\citename{Baumgartner \bgroup et al.\egroup
  }2008]{baumgartner2008decline}
Frank~R Baumgartner, Suzanna~L De~Boef, and Amber~E Boydstun.
\newblock 2008.
\newblock {\em The decline of the death penalty and the discovery of
  innocence}.
\newblock Cambridge University Press.

\bibitem[\protect\citename{Berinsky and Kinder}2006]{berinsky2006making}
Adam~J Berinsky and Donald~R Kinder.
\newblock 2006.
\newblock Making sense of issues through media frames: Understanding the
  {K}osovo crisis.
\newblock {\em The Journal of Politics}, 68(3):640--656.

\bibitem[\protect\citename{Budak \bgroup et al.\egroup }2016]{budak2016fair}
Ceren Budak, Sharad Goel, and Justin~M Rao.
\newblock 2016.
\newblock Fair and balanced? quantifying media bias through crowdsourced
  content analysis.
\newblock {\em Public Opinion Quarterly}, 80(S1):250--271.

\bibitem[\protect\citename{Card \bgroup et al.\egroup }2015]{card2015media}
Dallas Card, Amber Boydstun, Justin~H Gross, Philip Resnik, and Noah~A Smith.
\newblock 2015.
\newblock The {M}edia {F}rames {R}orpus: Annotations of frames across issues.
\newblock In {\em Proceedings of the 53rd Annual Meeting of the Association for
  Computational Linguistics and the 7th International Joint Conference on
  Natural Language Processing (Volume 2: Short Papers)}, pages 438--444.

\bibitem[\protect\citename{Card \bgroup et al.\egroup }2016]{card2016analyzing}
Dallas Card, Justin~H Gross, Amber Boydstun, and Noah~A Smith.
\newblock 2016.
\newblock Analyzing framing through the casts of characters in the news.
\newblock In {\em Proceedings of the 2016 Conference on Empirical Methods in
  Natural Language Processing}, pages 1410--1420.

\bibitem[\protect\citename{Cer \bgroup et al.\egroup }2018]{cer2018universal}
Daniel Cer, Yinfei Yang, Sheng-yi Kong, Nan Hua, Nicole Limtiaco, Rhomni~St
  John, Noah Constant, Mario Guajardo-Cespedes, Steve Yuan, Chris Tar, et~al.
\newblock 2018.
\newblock Universal sentence encoder.
\newblock {\em arXiv preprint arXiv:1803.11175}.

\bibitem[\protect\citename{Chen \bgroup et al.\egroup }2018]{chen2018learning}
Wei-Fan Chen, Henning Wachsmuth, Khalid Al~Khatib, and Benno Stein.
\newblock 2018.
\newblock Learning to flip the bias of news headlines.
\newblock In {\em Proceedings of the 11th International Conference on Natural
  Language Generation}, pages 79--88.

\bibitem[\protect\citename{Cohan \bgroup et al.\egroup
  }2019]{cohan2019pretrained}
Arman Cohan, Iz~Beltagy, Daniel King, Bhavana Dalvi, and Daniel~S Weld.
\newblock 2019.
\newblock Pretrained language models for sequential sentence classification.
\newblock {\em arXiv preprint arXiv:1909.04054}.

\bibitem[\protect\citename{Dernoncourt and Lee}2017]{dernoncourt2017pubmed}
Franck Dernoncourt and Ji~Young Lee.
\newblock 2017.
\newblock Pubmed 200k rct: a dataset for sequential sentence classification in
  medical abstracts.
\newblock {\em arXiv preprint arXiv:1710.06071}.

\bibitem[\protect\citename{Devlin \bgroup et al.\egroup }2018]{devlin2018bert}
Jacob Devlin, Ming-Wei Chang, Kenton Lee, and Kristina Toutanova.
\newblock 2018.
\newblock {BERT}: Pre-training of deep bidirectional transformers for language
  understanding.
\newblock {\em arXiv preprint arXiv:1810.04805}.

\bibitem[\protect\citename{Entman}1993]{entman1993framing}
Robert~M Entman.
\newblock 1993.
\newblock Framing: Toward clarification of a fractured paradigm.
\newblock {\em Journal of communication}, 43(4):51--58.

\bibitem[\protect\citename{Fan \bgroup et al.\egroup }2019]{basil2019}
Lisa Fan, Marshall White, Eva Sharma, Ruisi Su, Prafulla~Kumar Choubey, Ruihong
  Huang, and Lu~Wang.
\newblock 2019.
\newblock In plain sight: Media bias through the lens of factual reporting.
\newblock In {\em Proceedings of the 2019 Conference on Empirical Methods in
  Natural Language Processing and the 9th International Joint Conference on
  Natural Language Processing (EMNLP-IJCNLP)}, pages 6344--6350, Hong Kong,
  China, November. Association for Computational Linguistics.

\bibitem[\protect\citename{Field \bgroup et al.\egroup }2018]{field2018framing}
Anjalie Field, Doron Kliger, Shuly Wintner, Jennifer Pan, Dan Jurafsky, and
  Yulia Tsvetkov.
\newblock 2018.
\newblock Framing and agenda-setting in {R}ussian news: a computational
  analysis of intricate political strategies.
\newblock In {\em Proceedings of the 2018 Conference on Empirical Methods in
  Natural Language Processing}, pages 3570--3580.

\bibitem[\protect\citename{Fulgoni \bgroup et al.\egroup
  }2016]{fulgoni2016empirical}
Dean Fulgoni, Jordan Carpenter, Lyle Ungar, and Daniel Preotiuc-Pietro.
\newblock 2016.
\newblock An empirical exploration of moral foundations theory in partisan news
  sources.
\newblock In Nicoletta Calzolari~(Conference Chair), Khalid Choukri, Thierry
  Declerck, Sara Goggi, Marko Grobelnik, Bente Maegaard, Joseph Mariani, Helene
  Mazo, Asuncion Moreno, Jan Odijk, and Stelios Piperidis, editors, {\em
  Proceedings of the Tenth International Conference on Language Resources and
  Evaluation}, pages 3730--3736, may.

\bibitem[\protect\citename{Gentzkow and Shapiro}2010]{gentzkow2010drives}
Matthew Gentzkow and Jesse~M Shapiro.
\newblock 2010.
\newblock What drives media slant? evidence from us daily newspapers.
\newblock {\em Econometrica}, 78(1):35--71.

\bibitem[\protect\citename{Greene and Resnik}2009]{greene2009more}
Stephan Greene and Philip Resnik.
\newblock 2009.
\newblock More than words: Syntactic packaging and implicit sentiment.
\newblock In {\em Proceedings of human language technologies: The 2009 annual
  conference of the north american chapter of the association for computational
  linguistics}, pages 503--511.

\bibitem[\protect\citename{Gururangan \bgroup et al.\egroup
  }2020]{gururangan2020don}
Suchin Gururangan, Ana Marasovi{\'c}, Swabha Swayamdipta, Kyle Lo, Iz~Beltagy,
  Doug Downey, and Noah~A Smith.
\newblock 2020.
\newblock Don't stop pretraining: Adapt language models to domains and tasks.
\newblock {\em arXiv preprint arXiv:2004.10964}.

\bibitem[\protect\citename{Hube and Fetahu}2019]{hube2019neural}
Christoph Hube and Besnik Fetahu.
\newblock 2019.
\newblock Neural based statement classification for biased language.
\newblock In {\em Proceedings of the twelfth ACM international conference on
  web search and data mining}, pages 195--203.

\bibitem[\protect\citename{Iyyer \bgroup et al.\egroup
  }2014]{iyyer2014political}
Mohit Iyyer, Peter Enns, Jordan Boyd-Graber, and Philip Resnik.
\newblock 2014.
\newblock Political ideology detection using recursive neural networks.
\newblock In {\em Proceedings of the 52nd Annual Meeting of the Association for
  Computational Linguistics (Volume 1: Long Papers)}, volume~1, pages
  1113--1122.

\bibitem[\protect\citename{Jin and Szolovits}2018]{jin2018hierarchical}
Di~Jin and Peter Szolovits.
\newblock 2018.
\newblock Hierarchical neural networks for sequential sentence classification
  in medical scientific abstracts.
\newblock {\em arXiv preprint arXiv:1808.06161}.

\bibitem[\protect\citename{Liu \bgroup et al.\egroup }2019a]{liu2019detecting}
Siyi Liu, Lei Guo, Kate Mays, Margrit Betke, and Derry~Tanti Wijaya.
\newblock 2019a.
\newblock Detecting frames in news headlines and its application to analyzing
  news framing trends surrounding {US} gun violence.
\newblock In {\em Proceedings of the 23rd Conference on Computational Natural
  Language Learning (CoNLL)}, pages 504--514.

\bibitem[\protect\citename{Liu \bgroup et al.\egroup }2019b]{liu2019roberta}
Yinhan Liu, Myle Ott, Naman Goyal, Jingfei Du, Mandar Joshi, Danqi Chen, Omer
  Levy, Mike Lewis, Luke Zettlemoyer, and Veselin Stoyanov.
\newblock 2019b.
\newblock Ro{BERT}a: A robustly optimized {BERT} pretraining approach.
\newblock {\em arXiv preprint arXiv:1907.11692}.

\bibitem[\protect\citename{Niculae \bgroup et al.\egroup
  }2015]{niculae2015quotus}
Vlad Niculae, Caroline Suen, Justine Zhang, Cristian Danescu-Niculescu-Mizil,
  and Jure Leskovec.
\newblock 2015.
\newblock Quotus: The structure of political media coverage as revealed by
  quoting patterns.
\newblock In {\em Proceedings of the 24th International Conference on World
  Wide Web}, pages 798--808.

\bibitem[\protect\citename{Pang \bgroup et al.\egroup }2008]{pang2008opinion}
Bo~Pang, Lillian Lee, et~al.
\newblock 2008.
\newblock Opinion mining and sentiment analysis.
\newblock {\em Foundations and Trends{\textregistered} in Information
  Retrieval}, 2(1--2):1--135.

\bibitem[\protect\citename{Papalampidi \bgroup et al.\egroup
  }2019]{papalampidi2019movie}
Pinelopi Papalampidi, Frank Keller, and Mirella Lapata.
\newblock 2019.
\newblock Movie plot analysis via turning point identification.
\newblock {\em arXiv preprint arXiv:1908.10328}.

\bibitem[\protect\citename{Pappagari \bgroup et al.\egroup
  }2019]{pappagari2019hierarchical}
Raghavendra Pappagari, Piotr Zelasko, Jes{\'u}s Villalba, Yishay Carmiel, and
  Najim Dehak.
\newblock 2019.
\newblock Hierarchical transformers for long document classification.
\newblock In {\em 2019 IEEE Automatic Speech Recognition and Understanding
  Workshop (ASRU)}, pages 838--844. IEEE.

\bibitem[\protect\citename{Recasens \bgroup et al.\egroup
  }2013]{recasens2013linguistic}
Marta Recasens, Cristian Danescu-Niculescu-Mizil, and Dan Jurafsky.
\newblock 2013.
\newblock Linguistic models for analyzing and detecting biased language.
\newblock In {\em Proceedings of the 51st Annual Meeting of the Association for
  Computational Linguistics (Volume 1: Long Papers)}, pages 1650--1659.

\bibitem[\protect\citename{Reimers and Gurevych}2019]{reimers2019sentence}
Nils Reimers and Iryna Gurevych.
\newblock 2019.
\newblock Sentence-{BERT}: Sentence embeddings using {S}iamese {BERT}-networks.
\newblock {\em arXiv preprint arXiv:1908.10084}.

\bibitem[\protect\citename{Tsur \bgroup et al.\egroup }2015]{tsur2015frame}
Oren Tsur, Dan Calacci, and David Lazer.
\newblock 2015.
\newblock A frame of mind: Using statistical models for detection of framing
  and agenda setting campaigns.
\newblock In {\em Proceedings of the 53rd Annual Meeting of the Association for
  Computational Linguistics and the 7th International Joint Conference on
  Natural Language Processing (Volume 1: Long Papers)}, volume~1, pages
  1629--1638. ACL.

\bibitem[\protect\citename{van~den Berg \bgroup et al.\egroup
  }2020]{van2020doctor}
Esther van~den Berg, Katharina Korfhage, Josef Ruppenhofer, Michael Wiegand,
  and Katja Markert.
\newblock 2020.
\newblock Doctor who? {F}raming through names and titles in german.
\newblock In {\em Proceedings of The 12th Language Resources and Evaluation
  Conference}, pages 4924--4932, Marseille, France, May. European Language
  Resources Association.

\bibitem[\protect\citename{Wiebe \bgroup et al.\egroup
  }2004]{wiebe2004learning}
Janyce Wiebe, Theresa Wilson, Rebecca Bruce, Matthew Bell, and Melanie Martin.
\newblock 2004.
\newblock Learning subjective language.
\newblock {\em Computational linguistics}, 30(3):277--308.

\bibitem[\protect\citename{Wilson \bgroup et al.\egroup
  }2005]{wilson2005recognizing}
Theresa Wilson, Janyce Wiebe, and Paul Hoffmann.
\newblock 2005.
\newblock Recognizing contextual polarity in phrase-level sentiment analysis.
\newblock In {\em Proceedings of human language technology conference and
  conference on empirical methods in natural language processing}, pages
  347--354.

\bibitem[\protect\citename{Yano \bgroup et al.\egroup }2010]{yano2010shedding}
Tae Yano, Philip Resnik, and Noah~A Smith.
\newblock 2010.
\newblock Shedding (a thousand points of) light on biased language.
\newblock In {\em Proceedings of the NAACL HLT 2010 Workshop on Creating Speech
  and Language Data with Amazon’s Mechanical Turk}, pages 152--158.

\end{thebibliography}

\appendix
\section{Training details} 
\label{sec:appendixA}
For finetuning of BERT, we train for 10 epochs at a learning rate of 2e-5 with batch size 16, keeping the best model of these epochs on development data.

For finetuning of Roberta, we train for 10 epochs at a learning rate of 1e-5 with batch size 16, keeping the best model of these epochs on development data. 

For training WinSSC, we train for 10 epochs at a learning rate of 1.5e-5 with batch size 4, keeping the best model of these epochs on development data. 

For training the CIM models, we first generate embeddings.
We experiment with USE embeddings \cite{cer2018universal}, SBert embeddings \cite{reimers2019sentence}, RoBERTa's pooled output, the average of token embeddings from RoBERTa's final hidden layer, and the average across the CLS-token embeddings in the last 4 layers. 
We find the last type of input to perform the best. 
Next, we train the BiLSTM for 150 epochs with learning rate 1e-3 and batch size 32, with a hidden layer size of 1200, and 2 bi-directional layers. 

%We experimented with an ensemble approach which has proven effective in some other settings.
%For finetuning of SSC with adapted versions of Roberta, we train for 10 epochs at a learning rate of 1.5e-5 with batch size 6, keeping the best model of these epochs on development data. 

%\section{Examples} 
%\label{sec:appendixB}
%\input{sections/appendixB}

\end{document}